\documentclass[runningheads]{llncs}
\usepackage[T1]{fontenc}
\usepackage{framed,multirow}
\usepackage{amssymb}
\usepackage{latexsym}
\usepackage{booktabs}
\usepackage{xcolor}
\usepackage{amsmath}

\usepackage{graphicx}

\usepackage{lipsum} 
\usepackage[bottom]{footmisc}

\begin{document}
\title{HecVL: Hierarchical Video-Language Pretraining for Zero-shot Surgical Phase Recognition}
\titlerunning{HecVL}
%
\author{Kun Yuan\inst{1,3} \and
Vinkle Srivastav \inst{1,2} \and
Nassir Navab\inst{3} \and
Nicolas Padoy\inst{1,2}
}

\institute{University of Strasbourg, CNRS, INSERM, ICube, UMR7357, Strasbourg, France \and
IHU Strasbourg, Strasbourg, France \and CAMP, Technische Universit\"at M\"unchen, Munich, Germany
}
\maketitle 

\begin{abstract}
Natural language could play an important role in developing generalist surgical models by providing a broad source of supervision from raw texts. This flexible form of supervision can enable the model's transferability across datasets and tasks as natural language can be used to reference learned visual concepts or describe new ones. In this work, we present \emph{HecVL}, a novel hierarchical video-language pretraining approach for building a generalist surgical model. Specifically, we construct a hierarchical video-text paired dataset by pairing the surgical lecture video with three hierarchical levels of texts: at clip-level, atomic actions using transcribed audio texts; at phase-level, conceptual text summaries; and at video-level, overall abstract text of the surgical procedure. Then, we propose a novel fine-to-coarse contrastive learning framework that learns separate embedding spaces for the three video-text hierarchies using a single model. By disentangling embedding spaces of different hierarchical levels, the learned multi-modal representations encode short-term and long-term surgical concepts in the same model. Thanks to the injected textual semantics, we demonstrate that the \emph{HecVL} approach can enable zero-shot surgical phase recognition without any human annotation. Furthermore, we show that the same HecVL model for surgical phase recognition can be transferred across different surgical procedures and medical centers. The code is available at \href{https://github.com/CAMMA-public/SurgVLP}{https://github.com/CAMMA-public/SurgVLP}.

\end{abstract}
\section{Introduction}

\footnote{\small \textit{This manuscript has been accepted for publication and will be included in the proceedings of MICCAI 2024.}}

\begin{figure*}
  \centering
  \includegraphics[width=\columnwidth]{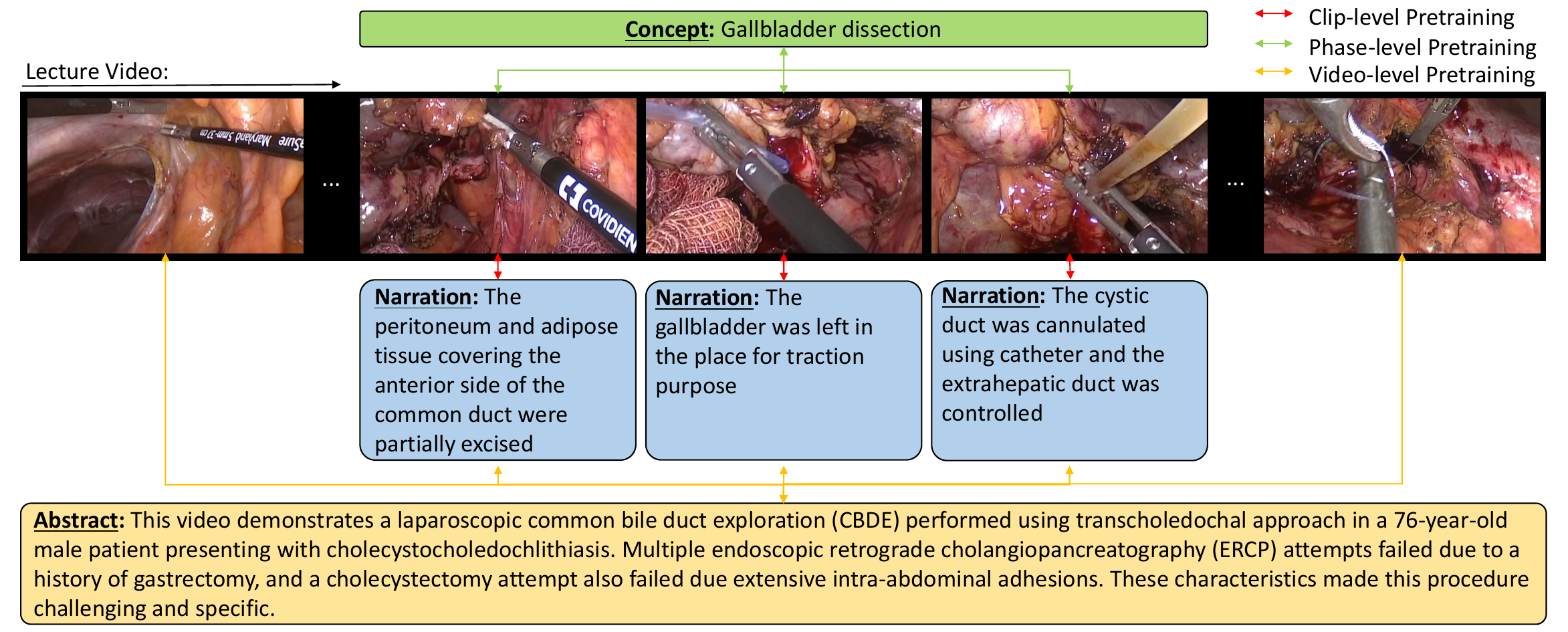} 
  \caption{Hierarchical video-text pairs in surgical lecture videos. Conventional methods~\cite{yuan2023learning} utilize only clip-level video-text pairs, while our HecVL utilizes different hierarchical levels of pairs to perform video-language pretraining.}
  \label{fig1}
\end{figure*}

Developing a single neural network model capable of adapting to different datasets and tasks stands as a key objective for computer vision. Recent breakthroughs in computer vision methods have begun to fulfill this goal by transitioning from task-specific models~\cite{he2016deep,chen2017deeplab,he2017mask} to generalist models~\cite{zou2023generalized,radford2021learning}. These generalist models have shown potential in solving a wide range of downstream tasks and datasets, including various types of object segmentation~\cite{zou2024segment} and zero-shot image and video classification~\cite{lin2023match}. An essential feature of these models is their ability to be supervised through natural language texts. The generality of natural language allows it to express a broader set of visual concepts, thereby effectively guiding and supervising these models~\cite{chen2022unified}.

Yet, within the domain of surgical video analysis, predominant methods still lean toward task-specific models~\cite{twinanda2016endonet,nwoye2021rendezvous,wu2021multi}. This is mainly due to the inherent complexity present in the surgical videos, i.e., surgical videos can last several hours while capturing intricate hierarchical surgical activities. Therefore, those methods manually define different levels of categories and annotate large amounts of frames to provide extensive supervision. However, the procedure- and center-specific annotations lead to degraded transferability across procedures and medical centers~\cite{lavanchy2023challenges}. While a surgical foundation model~\cite{wang2023foundation} is proposed to address the above issue, it focuses only on pure images and ignores the complementary information from other modalities, i.e., language. Also, it still requires finetuning on the downstream dataset to enable transferability. Considering that natural language texts have become a unifying element for generalist models, this work explores whether they can be used to both understand the hierarchical intricacies of surgical videos and enable the generalized zero-shot transfer by processing category labels into texts, without the need for manual annotation. As the task of surgical phase recognition is essential for computer-assisted surgery~\cite{padoy2012statistical,blum2010modeling,twinanda2016endonet,jin2017sv,czempiel2020tecno}, we use it as a suitable test bench to evaluate our joint visual and textual hierarchical representations.

This work introduces \emph{HecVL}, a Hierarchical Encoded Contrastive Video-language pretraining framework, which learns rich multi-modal representations at different hierarchies of surgical video. Developing such an approach presents a significant challenge due to the lack of surgical video datasets with hierarchical textual supervision. SurgVLP~\cite{yuan2023learning} has introduced the first large-scale video-text paired dataset, i.e., \emph{SVL}, by transcribing hundreds of surgical lecture videos into narration texts. We extend \emph{SVL} dataset by incorporating hierarchical-level texts using the metadata of each lecture video. We construct three levels of the hierarchical video-text pairs for each surgical lecture video: \emph{clip-level}, \emph{phase-level}, and \emph{video-level}. The clip-level video-text pairs contain short video clips of few seconds duration along with narration texts transcribed from lecture audio for capturing the short-term activity. The phase-level video-text pairs contain longer video segments with conceptual text summaries for capturing longer surgical video activity. Finally, the video-level video-text pairs are the entire surgical lecture videos paired with abstract paragraphs encapsulating the goal and the main key points of the surgery. These three levels of hierarchical video-text pairs allow for a more detailed understanding of surgical procedures, capturing both the atomic details and broader contexts, as illustrated in Fig.~\ref{fig1}.

Given the hierarchical video-text pair dataset, we propose a \emph{fine-to-coarse contrastive} learning strategy to effectively exploit the hierarchical textual information encoded in the dataset. We construct three separate embedding spaces for each type of hierarchical video-text pair. We first build up a fine-grained embedding space using clip-level video-text pairs, followed by aggregating the fine-grained features to construct the coarse-grained embedding spaces, which embed phase-level and video-level text pairs. We learn these three different embedding spaces through multi-modal contrastive learning using the InfoNCE loss~\cite{oord2018representation}. We show in the experiments that our fine-to-coarse contrastive learning strategy outperforms the approach of projecting all hierarchical texts into a single embedding space and learning only one such space.

We demonstrate the zero-shot transferability and the generalization of our approach by performing surgical phase recognition on three different surgical procedures, cholecystectomy~\cite{twinanda2016endonet}, hysterectomy~\cite{wang2022autolaparo}, and gastric bypass~\cite{lavanchy2023challenges}, without using any ground truth labels. The learned multi-modal representations demonstrate their transferability in not only identifying surgical concepts across various surgical procedures but also in extending to different medical centers. We hope that the \emph{HecVL} approach could pave the path for developing more generalist models in the domain of surgical computer vision.

\section{Method}
\label{sec:method}

\begin{figure*}[!htb]
  \centering
  \includegraphics[width=0.98\columnwidth]{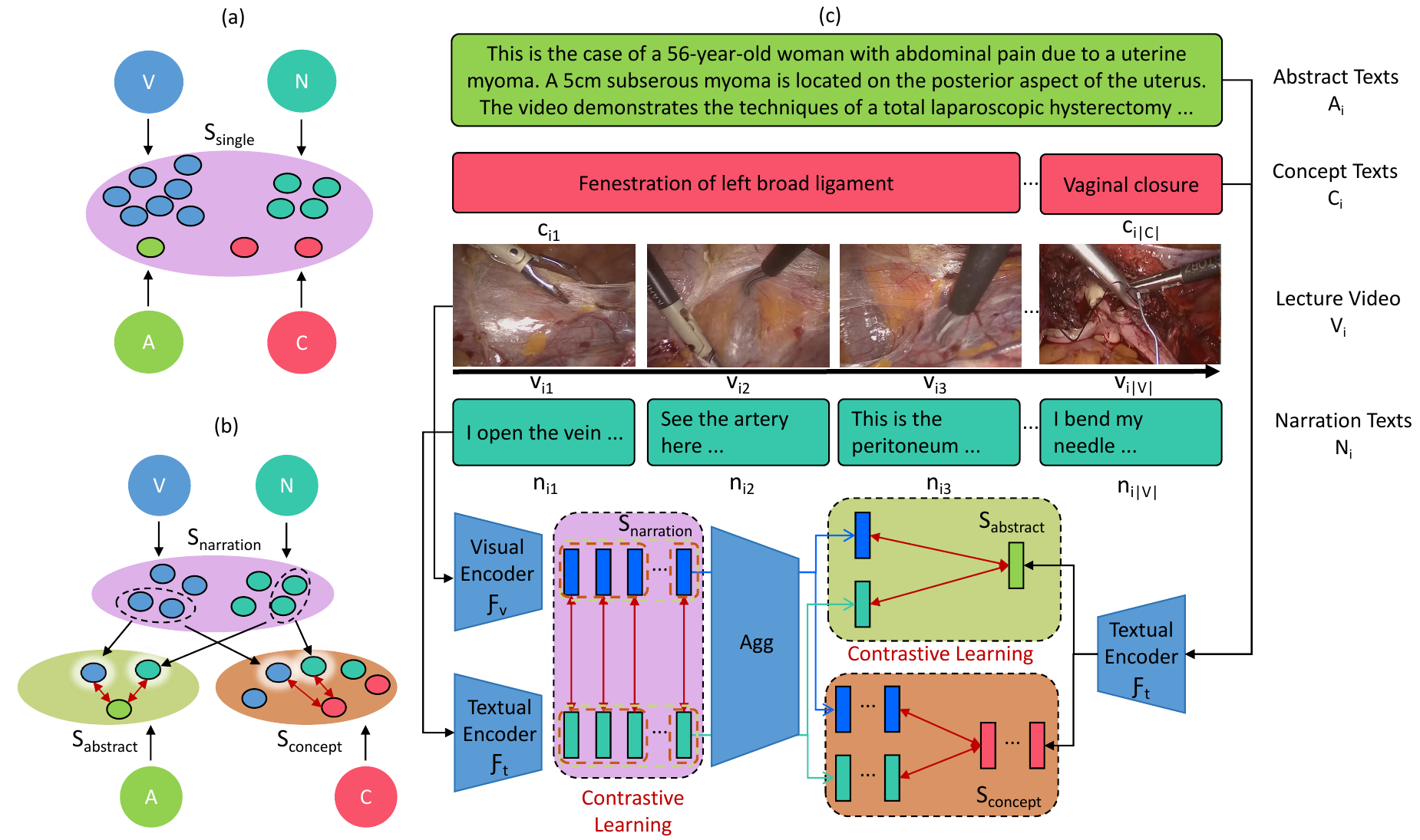} 
  \caption{Pipeline of the \emph{HecVL} approach. (a) Conventional video-language methods embed video clips and texts of different granularities into the same embedding space. (b) The \emph{HecVL} approach considers the granularity differences and constructs three embedding spaces for clip-, phase-, and video-level representation learning. (c) The fine-grained embedding space ($S_{narration}$) is learned first, followed by learning of coarse-space embedding spaces ($S_{abstract}$ and $S_{concept}$) using a temporal aggregation function to aggregate the visual and the textual embeddings.}
  \label{fig2}
\end{figure*}

We propose \emph{HecVL}, a novel hierarchical video-language pretraining method that learns multi-modal embeddings by capturing clip-, phase-, and video-level video-text pairs from surgical lecture videos. Fig.~\ref{fig2} gives an overview of our method. Sec.~\ref{hier_annotation} describes the construction of hierarchical video-text pairs. Sec.~\ref{hier_pretrain} formalizes the fine-to-coarse contrastive learning strategy. Sec.~\ref{learning_objectives} and~\ref{train_pipeline} describe the loss function and the training pipeline, respectively.

\subsection{Hierarchical video-text pairs}
\label{hier_annotation}

The \emph{HecVL} approach is designed to leverage a hierarchically annotated video-text pair dataset, $D = \{ (V_i, N_i, C_i, A_i) \}_{i=1}^{|D|}$, where $V_i$ is a long surgical lecture video composed of a sequence of short-term video clips (each lasting tens of seconds). Each lecture video $V_i$ is paired with three levels of textual annotations from different levels of granularities ranging from fine-grained to coarse-grained, i.e., \textbf{clip-level} narration texts ($N_i$), \textbf{phase-level} concept texts ($C_i$), and \textbf{video-level} abstract texts ($A_i$). 

The clip-level narration texts ($N_i$) are sequences of narrations describing the atomic actions for short-term video clips, the phase-level concept texts ($C_i$) are sequences of conceptual text summaries describing the high-level surgical activities for long-term video phases, and the video-level abstract paragraph texts ($A_i$) are the abstract paragraph texts summarizing the entire surgical lecture video including patient's history and surgical technique. These three levels of video-text pairs provide complementary textual supervision at multiple hierarchies for representation learning, as illustrated in Fig.~\ref{fig1}.

\subsection{Fine-to-coarse contrastive learning}
\label{hier_pretrain}

Given the hierarchically annotated video-text pair dataset as described above, we aim to optimize a visual encoder $\mathcal{F}_v$ and a textual encoder $\mathcal{F}_t$ for the multi-modal hierarchical representation learning. This is achieved by constructing different embedding spaces for hierarchical video-text pairs, as described below. 
\subsubsection{Embedding spaces at different hierarchical levels:} given the visual encoder $\mathcal{F}_v$ and the textual encoder $\mathcal{F}_t$, we first extract \textbf{clip-level} visual embeddings $\mathcal{F}_v(v_{ij})$ and textual embeddings $\mathcal{F}_t(n_{ij})$ from the short-term video clips $v_{ij} \in V_i$ and their corresponding clip-level narration texts $n_{ij} \in N_i$. These multi-modal embeddings are represented in the fine-grained embedding space $S_{narration}$. 

Then, we construct another embedding space $S_{concept}$ by exploiting \textbf{phase-level} textual supervision. We define $V^c$ and $N^c$ as the sets of short-term video clips $v_{ij}$ and narration texts $n_{ij}$ temporally corresponding to phase-level concept texts $c_{ij} \in C_i$. We extract the textual embeddings $\mathcal{F}_t(c_{ij})$ using the textual encoder $\mathcal{F}_t$. Subsequently, we define an aggregator function $Agg()$, which takes clip-level visual and textual embeddings, $V^c$ and $N^c$, as input and performs average pooling on them. The aggregated visual embeddings  $Agg(\mathcal{F}_v(V^c))$ and textual embeddings $Agg(\mathcal{F}_t(N^c))$ are represented in embedding space $S_{concept}$.

Finally, we construct a \textbf{video-level} embedding space, $S_{abstract}$, using video-level abstract texts ($A_i$). Similar to constructing the phase-level embedding space, we define $V^a$ and $N^a$ as the evenly sampled sets of short-term video clips $v_{ij}$ and narration texts $n_{ij}$. As there is only one abstract text for the entire video, $V^a$ and $N^a$ correspond to the video-level abstract texts $A_i$. We extract the textual embeddings $\mathcal{F}_t(A_i)$ from the abstract text $A_i$. Then, the aggregator function $Agg()$ is employed to aggregate the clip-level visual and textual embeddings. The resulting aggregated visual embeddings $Agg(\mathcal{F}_v(V^a))$ and textual embeddings $Agg(\mathcal{F}_t(N^a))$ are represented in space $S_{abstract}$. The construction of all three embedding spaces is illustrated in the Fig.~\ref{fig2}.

Fig.~\ref{fig2} (a) shows another way to conduct video-language pretraining by projecting all the video clips and the three levels of texts to a single embedding space. We show in experiments that this increases the ambiguity as video clips might be pushed to both narration and concept texts with dissimilar semantics.

\subsection{Training objectives}
\label{learning_objectives}

We propose a joint contrastive loss function to enhance the similarity score between matching visual and textual embeddings compared to non-matching pairs in $S_{narration}$, $S_{concept}$, and $S_{abstract}$. At the clip level, we use the loss function from the SurgVLP~\cite{yuan2023learning} $\mathcal{L}_{clip}$ (also given in the supplementary) to correlate short-term video clips with the narrations from two different automatic speech recognition (ASR) systems. At the phase- and the video-level, we use the InfoNCE~\cite{oord2018representation} loss to correlate the aggregated short-term visual and textual embeddings to phase- and video-level textual embeddings, respectively, as given below:
\begin{equation}
\scalebox{0.69}{
    \begin{math}
    \begin{aligned}
        \mathcal{L}_{phase} = & - \frac{1}{B} \sum_{i=1}^{B} \log \left( \frac{\exp(Agg(\mathcal{F}_v(V^c))^{T} \cdot \mathcal{F}_t(c_{i})/\tau)}{\sum^B_{j=1} \exp(Agg(\mathcal{F}_v(V^c))^{T} \cdot \mathcal{F}_t(c_{j})/\tau)} + \frac{\exp(Agg(\mathcal{F}_t(N^c))^{T} \cdot \mathcal{F}_t(c_{i})/\tau)}{\sum^B_{j=1} \exp(Agg(\mathcal{F}_t(N^c))^{T} \cdot \mathcal{F}_t(c_{j})/\tau)} \right)
    \end{aligned}
    \end{math}
}
\label{phase_loss}
\end{equation}

\begin{equation}
\scalebox{0.69}{
    \begin{math}
    \begin{aligned}
        \mathcal{L}_{video} = & - \frac{1}{B} \sum_{i=1}^{B} \log \left( \frac{\exp(Agg(\mathcal{F}_v(V^A))^{T} \cdot \mathcal{F}_t(A_{i})/\tau)}{\sum^B_{j=1} \exp(Agg(\mathcal{F}_v(V^A))^{T} \cdot \mathcal{F}_t(A_{j})/\tau)} + \frac{\exp(Agg(\mathcal{F}_t(N^A))^{T} \cdot \mathcal{F}_t(A_{i})/\tau)}{\sum^B_{j=1} \exp(Agg(\mathcal{F}_t(N^A))^{T} \cdot \mathcal{F}_t(A_{j})/\tau)} \right)
    \end{aligned}
    \end{math}
}
\label{video_loss}
\end{equation}

Here, $B$ is the batch size, and $\tau$ is a temperature hyper-parameter, which regulates the probability distribution over positive and negative pairs within the embedding space. The numerator term denotes the cosine similarity between matched visual and the textual pairs, i.e., \emph{positive pairs}, and the denominator term denotes the cosine similarity between the unmatched visual and the textual pairs, i.e., \emph{negative pairs}.

\subsection{Training pipeline}
\label{train_pipeline}

Given the previously described training objectives, the challenge lies in effectively training $\mathcal{F}_v$ and $\mathcal{F}_t$ across all three levels of embedding spaces. We aim to train only one set of visual and textual encoders for all three levels of embedding spaces, ensuring the encoders are optimized for capturing both short-term and long-term semantics. We propose an \emph{alternating training strategy}, i.e., we first optimize $L_{clip}$ for $m$ batches; subsequently, we optimize $L_{phase}$ for $n$ batches and $L_{video}$ for $l$ batches, and then we repeat. We observe that the proposed training strategy not only converges faster but also circumvents the catastrophic forgetting issue~\cite{goodfellow2013empirical} that could arise when training the model on clip-level embeddings and then fine-tuning it for phase- and video-level embeddings, or vice-versa.

\section{Experiment setup}
\label{sec:experiments_setup}
\subsection{Dataset}
\textbf{Pretraining dataset:} we use the surgical lecture videos from the Surgical Video Lecture dataset (SVL) for the pretraining, which was proposed by the SurgVLP \cite{yuan2023learning}. We further expand the dataset by including additional phase- and video-level video-text pairs using the metadata of each lecture video. The metadata for each lecture video contains the title of the procedure, the abstract summary, and the key steps. In total, we have $25,578$ clip-level, $10,304$ phase-level, and $1,076$ video-level video-text pairs.

\noindent\textbf{Downstream datasets and evaluation:} we perform the evaluation on three public datasets: Cholec80~\cite{twinanda2016endonet}, AutoLaparo~\cite{wang2022autolaparo}, StrasBypass70 and BernBypass70~\cite{lavanchy2023challenges}. In our evaluation, we perform surgical phase recognition in the zero-shot setting, which directly evaluates the model on downstream datasets without performing any fine-tuning. Here, class labels are transformed into textual prompts, and their embeddings categorize the image visual embeddings, reflecting the joint embedding space's effectiveness. (details of the constructed
textual prompts are given in the Supplementary).

\subsection{Implementation details}

\textbf{Network architecture:}  We use the ResNet-50 model~\cite{he2016deep} pretrained on ImageNet as visual encoder $\mathcal{F}_v$ and BioClinicalBert~\cite{huang2019clinicalbert} as the textual encoder $\mathcal{F}_t$. We sample $4$, $8$, and $32$ frames for each clip-level, phase-level, and video-level video segment. We encode the frames, followed by an average pooling to generate a feature vector for a video segment. The architectures of visual and text encoders, $\mathcal{F}_v$ and $\mathcal{F}_t$, are the same as SurgVLP~\cite{yuan2023learning} for a fair comparison. 

\noindent\textbf{Training parameters:} we pretrain the model with one $80$ GB NVIDIA A$100$ GPUs for $200$ epochs. We use the AdamW~\cite{loshchilov2017decoupled} optimizer with a learning rate of $5e-5$. We alternatively train with $m=25$ batches of clip-level pairs, followed by $n=15$ and $l=115$ batches of phase- and video-level pairs. We use a batch size $B$ of $120/60/10$ per GPU for clip-/phase-/video-level video-text pairs.

\section{Results and discussions}

\begin{table*}[t!]
\centering
\caption{Zero-shot phase recognition results on cholecystectomy and hysterectomy. CLIP**/CLIP*: random/OpenAI initialized CLIP model and pretraining with SVL dataset~\cite{yuan2023learning}.}
\label{tab:zero-shot}
    \begin{tabular}{@{}cccccc@{}}
    \toprule
    Model & Pretraining dataset & \multicolumn{2}{c}{Cholec-80} & \multicolumn{2}{c}{Autolaparo} \\
      & & Top-1 Acc. & F1 Score & Top-1 Acc. & F1 Score \\ \hline
    \midrule
    MIL-NCE~\cite{miech2020end} & Howto100M & 7.8 & 7.3 & 9.9 & 7.9 \\
    CLIP~\cite{radford2021learning} & CLIP400M & 30.8 & 13.1 & 17.4 & 9.1  \\
    \midrule
    CLIP**~\cite{radford2021learning} & SVL-Pretrain & 29.4 & 10.4 & 15.3 & 10.9 \\
    CLIP*~\cite{radford2021learning} & SVL-Pretrain & {33.8} & {19.6} & 18.9 & 16.2 \\
    SurgVLP~\cite{yuan2023learning} & SVL-Pretrain & {34.7} & {24.4} & {21.3} & {16.6} \\
    HecVL & SVL-Pretrain & \textbf{41.7}& \textbf{26.3} & \textbf{23.3} & \textbf{18.9} \\
    \hline
    \bottomrule
    
    \end{tabular}
\end{table*}

\subsection{Zero-shot phase recognition}

Results on zero-shot surgical phase recognition demonstrate if the learned joint visual and textual representations can correlate semantically similar surgical scene images and surgical texts. We compare our method to CLIP~\cite{radford2021learning} to show the benefits of surgical-specific pretraining, and to SurgVLP to emphasize hierarchical pretraining advantages. Tab. \ref{tab:zero-shot} and \ref{tab:zero-shot-bypass} show that our \emph{HecVL} achieves state-of-the-art performance for all the datasets in the zero-shot setting. The consistent boost across cholecystectomy~\cite{twinanda2016endonet}, hysterectomy~\cite{wang2022autolaparo}, and gastric bypass~\cite{lavanchy2023challenges} procedures show the generalizable and transferable features of \emph{HecVL} across different surgical types. Also, we show significant improvement compared to the methods pretrained on the conventional computer vision datasets, i.e., MIL-NCE~\cite{miech2020end}, CLIP~\cite{radford2021learning}, which fails in recognizing the surgical concepts.

\begin{table*}[t!]
\centering
\caption{Zero-shot phase recognition results across medical centers on gastric bypass surgery. We evaluate our model on the test split of StrasBypass70 and BernBypass70 from the University Hospital of Strasbourg and Bern University Hospital.}
\label{tab:zero-shot-bypass}
    \begin{tabular}{@{}cccccc@{}}
    \toprule
    Model & Pretraining dataset & \multicolumn{2}{c}{StrasBypass70} & \multicolumn{2}{c}{BernBypass70} \\
      &  & Top-1 Acc. & F1 Score & Top-1 Acc. & F1 Score \\ \hline
    \midrule
    MIL-NCE~\cite{miech2020end} & Howto100M & 3.8 & 2.7 & 2.5 & 2.0 \\
    CLIP~\cite{radford2021learning} & CLIP400M & 12.0 & 6.8 & 9.7 & 5.1  \\
    \midrule
    CLIP**~\cite{radford2021learning} & SVL-Pretrain & 6.3 & 3.5 & 4.9 & 2.3 \\
    CLIP*~\cite{radford2021learning} & SVL-Pretrain & 15.8 & 8.6 & 17.8 & 7.1 \\
    SurgVLP~\cite{yuan2023learning} & SVL-Pretrain & 10.8 & 6.9 & 11.4 & 7.2 \\
    HecVL & SVL-Pretrain & \textbf{26.9} & \textbf{18.3} & \textbf{22.8} & \textbf{13.6} \\
    \hline
    \bottomrule
    
    \end{tabular}
\end{table*}

\begin{table*}[t!]
\centering
\caption{Ablation study on different levels. We conduct zero-shot phase recognition and report F1 Score. Single: single embedding space as in Fig.~\ref{fig1}(a).}
\label{tab:ablation}
\begin{tabular}{@{}cccccccc@{}}
\toprule
Model & Action & Phase & Abstract & Cholec80 & AutoLaparo & StrasBypass70 & BernBypass70 \\
\midrule
SurgVLP  & \checkmark & $\times$ & $\times$ & 24.4 & 16.6 & 6.9 & 7.2\\
\emph{HecVL} & \checkmark & \checkmark & $\times$ & 25.2 & 16.9 & 15.5 & 12.9 \\
Single& \checkmark  & \checkmark & \checkmark & 25.6 & 15.7 & 14.6 & \textbf{14.0}\\
\emph{HecVL}  & \checkmark  & \checkmark & \checkmark & \textbf{26.3} & \textbf{18.9} & \textbf{18.3} & 13.6\\
\bottomrule
\end{tabular}
\end{table*}

\subsection{Multi-center phase recognition}
Here, we examine the ability of the \emph{HecVL} approach to transfer the knowledge learned from hierarchical video-text data to different medical centers, as shown in Tab.~\ref{tab:zero-shot-bypass}. Overall, our HecVL model achieves the best performance across two medical centers compared to the other methods. Interestingly, the performance of the BernBypass70 is lower than the StrasBypass70. This may be because there are significant differences in the workflow followed at the Bern center, with many phases and steps not routinely performed. Therefore, the textual prompts designed based on the Strasbourg center's protocol (see Supplementary for more details) lead to degraded performance when applied to the different centers. To address this, a center-specific textual prompts construction is required.

\subsection{Ablation study}
Tab. \ref{tab:ablation} provides the ablation analysis of the contribution of each level of video-text pairs. Specifically, adding the phase-level video-text pairs yields significant improvements in gastric bypass surgical phase recognition compared to the SurgVLP. This trend is pronounced in the zero-shot scenario across all surgical procedures. We also compare to a baseline model: Single in Fig.~\ref{fig2} (a), which embeds action, phase, and abstract texts in a single embedding space, to support our fine-to-coarse strategy. Tab. \ref{tab:ablation} shows that Single model has inconsistent performance across datasets. This inconsistency implies that the single embedding approach might blur essential distinctions across hierarchical levels compared to maintaining separate embedding spaces for different hierarchical levels.

\section{Conclusion}
The next generation of scalable and generalizable surgical computer vision systems demands multi-modality models capable of adapting to different surgical procedures with little to no manual annotations. In this work, we design \emph{HecVL}, a single multi-modality model \emph{HecVL} capable of adapting to the different surgical procedures and centers without using any manual annotations. The core of our contribution lies in developing a hierarchical contrastive learning strategy to exploit textual supervision at multiple granular levels, ranging from short-term surgical actions to long-term high-level surgical concepts. Extensive experimental results demonstrate its efficacy in achieving zero-shot surgical phase recognition across different procedures and medical centers. 

\section{Acknowledgments}
This work has received funding from the European Union (ERC, CompSURG, 101088553). Views and opinions expressed are however those of the authors only and do not necessarily reflect those of the European Union or the European Research Council. Neither the European Union nor the granting authority can be held responsible for them. This work was also partially supported by French state funds managed by the ANR under Grant ANR-10-IAHU-02. This work was granted access to the HPC resources of IDRIS under the allocations AD011013704R1, AD011011631R2, and AD011011631R3 made by GENCI. The authors would also like to acknowledge the High-Performance Computing Center of the University of Strasbourg for providing access to computing resources funded by the Equipex Equip@Meso project (Programme Investissements d’Avenir) and the CPER Alsacalcul/Big Data.

\bibliographystyle{splncs04}
\bibliography{mybibliography}

\end{document}